# Deep Learning-based Prediction of Stress and Strain Maps in Arterial Walls for Improved Cardiovascular Risk Assessment

Yasin Shokrollahi[1], Pengfei Dong[1], Xianqi Li[2*], Linxia Gu[1*]

[1] Department of Biomedical and Chemical Engineering and Sciences, Florida Institute of Technology, Melbourne, FL 32901, USA

[2] Department of Mathematical Sciences, Florida Institute of Technology, Melbourne, FL 32901, USA

* Corresponding author：xli@fit.edu, gul@fit.edu

**Abstract**

Conducting computational stress-strain analysis using finite element methods (FEM) is a common approach when dealing with the complex geometries of atherosclerosis, which is a leading cause of global mortality and complex cardiovascular disease. The considerable expense linked to FEM analysis encourages the substitution of FEM with a considerably faster data-driven machine learning (ML) approach. This study investigated the potential of end-to-end deep learning tools as a more effective substitute for FEM in predicting stress-strain fields within 2D cross sections of arterial wall. We first proposed a U-Net based fully convolutional neural network (CNN) to predict the von Mises stress and strain distribution based on the spatial arrangement of calcification within arterial wall cross-sections. Further, we developed a conditional generative adversarial network (cGAN) to enhance, particularly from the perceptual perspective, the prediction accuracy of stress and strain field maps for arterial walls with various calcification quantities and spatial configurations. On top of U-Net and cGAN, we also proposed their ensemble approaches, respectively, to further improve the prediction accuracy of field maps. Our dataset, consisting of input and output images, was generated by implementing boundary conditions and extracting stress-strain field maps. The trained U-Net models can accurately predict von Mises stress and strain fields, with structural similarity index scores (SSIM) of 0.854 and 0.830 and mean squared errors of 0.017 and 0.018 for stress and strain, respectively, on a reserved test set. Meanwhile, the cGAN models in a combination of ensemble and transfer learning techniques demonstrate high accuracy in predicting von Mises stress and strain fields, as evidenced by SSIM scores of 0.890 for stress and 0.803 for strain. Additionally, mean squared errors of 0.008 for stress and 0.017 for strain further support the model's performance on a designated test set. Overall, this study developed a surrogate model for finite element analysis, which can accurately and efficiently predict stress-strain fields of arterial walls regardless of complex geometries and boundary conditions.

# 1. Introduction

Atherosclerosis is a complex cardiovascular disease characterized by plaque accumulation in the arterial walls, leading to the narrowing and hardening of the arteries. It is a major cause of heart disease, stroke, and other cardiovascular diseases, making it one of the leading causes of death worldwide. The disease is caused by genetic, environmental, and lifestyle factors, including high cholesterol, high blood pressure, smoking, obesity, and diabetes (Libby et al., 2002; Weber and Noels, 2011). The mechanical properties of the plaque, such as its stiffness and strength, are important determinants of its stability and the risk of rupture. Therefore, predicting the stress-strain field maps of the plaque can provide valuable insights into its mechanical behavior and help identify regions that may be prone to rupture. Several studies have demonstrated the potential of computational models and imaging techniques for predicting the stress-strain field maps of atherosclerotic plaques and assessing their risk of rupture (Church and Miller, 2016; Krams et al., 1997). Cheng et al. (Cheng et al., 2014) found that maximum von Mises stress was significantly higher in ruptured plaques compared to stable plaques. Therefore, predicting the von Mises stress could be an essential tool for preventing plaque rupture and reducing the incidence of cardiovascular events. The FEM, known as the finite element method, is the conventional numerical technique employed for stress-strain analysis of structures. It revolves around solving partial differential equations to evaluate the system's behavior (Bathe, 2006; Lam et al., 2022; Reddy, 2019; Shokrollahi et al., 2022c). However, FEM simulations can be expensive, especially for highly nonlinear analyses or for complex geometries. Consequently, substantial endeavors have been devoted to substituting FEM with machine learning (ML) techniques, commonly employed for surrogate modeling (Bhaduri et al., 2018; Cristianini and Shawe-Taylor, 2000; Gholami et al., 2023; Mollaei Ardestani et al., 2023; Shokrollahi et al., 2022a; Shokrollahi et al., 2022b; Williams, 1998) of pertinent quantities of interest.

Madani et al.(Madani et al., 2019) presented a method for bridging FEM and ML for predicting the von Mises stress distribution in arterial walls affected by atherosclerosis. The authors utilized a FEM to simulate the mechanical behavior of arterial walls and generated a large dataset of stress values for varying degrees of plaque build-up. The dataset was then used to train an ML model that could predict stress values based on input parameters such as plaque thickness and the diameter of the arterial lumen. Liu et al.(Li et al., 2021) used a CNN model to automatically segment atherosclerotic plaques in intravascular ultrasound (IVUS) images. The authors used a dataset of IVUS images and achieved an accuracy of 90% in plaque segmentation. The segmentation results were then used to simulate the mechanical properties of the plaque using FEM. The study showed that the mechanical properties of the plaque were highly dependent on its morphology and composition. Chau et al.(Chau et al., 2004) performed Optical Coherence Tomography (OCT) imaging of atherosclerotic plaques in human cadaveric coronary arteries and used finite element analysis to assess the mechanical properties of the plaques. They found that the fibrous cap's thickness and the lipid core's size were important factors in determining the mechanical stability of the plaque. Plaques with thinner fibrous caps and larger lipid cores were found to have higher stress concentrations, which may contribute to plaque rupture. Cilla et al.(Cilla et al., 2012) presented a study on applying ML techniques to determine plaque vulnerability, an important factor in predicting the risk of heart attack and stroke. The authors discuss the limitations of traditional methods for assessing plaque vulnerability and the potential benefits of using ML techniques to analyze patient imaging data. The study involved analyzing patients with carotid artery disease, and the results show that ML techniques can effectively determine plaque vulnerability. The authors used conventional ML algorithms, such as support vector machines (SVM) and random forests (RF), to analyze imaging data and extract relevant features. They also developed a new "plaque volume score" method combining different features to predict plaque vulnerability better. The results show that ML algorithms can accurately predict plaque vulnerability, with an accuracy of up to 90%. The authors suggest that these techniques can be used in clinical practice to improve risk assessment and treatment planning for patients with carotid artery disease. While all these papers discuss predicting stress distribution, none of them address the issue of how well their predictions would hold up in the face of variability in arterial wall geometries, such as changes in calcification angle, thickness, and number, or shifts in lumen location and fibrous thickness. Our innovative approach considers the variability in calcification numbers while predicting stress and strain in arterial walls.

This study aims to employ deep learning models for predicting stress and strain distribution in arterial walls under blood pressure loading, specifically considering various spatial calcification distributions. Encoder-decoder networks, particularly U-Net networks (Ronneberger et al., 2015), have demonstrated effective image mapping due to their ability to capture high-resolution details and low-level features by propagating context information from lower to higher layers. Moreover, skip connections help prevent the vanishing gradient problem during model training. Conversely, cGAN, which learns a mapping from an observed image and random noise vector to an output image, is utilized for predicting stress-strain maps using a U-Net generator architecture and a PatchGAN discriminator, highlighting the superiority of cGAN in predicting strain maps of arterial walls compared to the U-Net model, particularly from the perceptual perspective (Ledig et al., 2017). This research implements deep learning models incorporating U-Net and cGAN, showcasing their generalization ability in predicting stress and strain maps of arterial walls. Moreover, we demonstrate the benefits of using ensemble and transfer learning strategies for improved accuracy. Impressively, our results display high accuracy even when trained on a mere 4,000 images, which we achieve through data augmentation grounded in the underlying physics of the issue. The structure of the manuscript is as follows: In Section 2, the proposed methodology is discussed. Section 3 showcases the prediction results and analysis, while Section 4 offers discussions and conclusions.

## 2. Materials and Methods
### 2.1. Generating Arterial Walls Model

Our training database comprises finite element simulations of different arterial geometries and boundary conditions. Our collaborators provided clinical data obtained through a two-step deep learning process for segmenting calcified coronary plaque in intravascular OCT images. This method involved combining a U-Net-based model with a GAN to identify calcified plaques in OCT images effectively. We utilized this segmentation data to construct a 3D model of a calcified artery based on OCT data. To streamline the process and generate various artery geometries rapidly, we first developed a 2D parametric model using Python scripting in Abaqus/Explicit software version 2019, drawing from the simplified clinical data and our observations. Python scripts in Abaqus can automate repetitive tasks, customize the software's behavior, and integrate it with other tools in a software pipeline. For example, a script could generate a series of similar simulations with varying parameters or post-process simulation results and generate customized visualizations. As shown in Fig. 1, we simplified the assumption that the cross-section consists of a series of circles and splines. The specifications of the circles, including a lumen, fibrous, artery, calcium deposits, and the number of calcium, were randomly varied according to Table 1. To produce calcification in a free-form shape, we began by selecting two circles as the inner and outer boundaries of the calcification, each with a random radius. We then randomly chose a calcification angle between 0 and 180 degrees. Next, we generated 10 points with equal intervals within this angle on the calcification's inner and outer boundaries. Finally, we utilized spline tools in Python to connect all these points and form a closed surface as the calcification. The number of calcifications may also be either one or two.

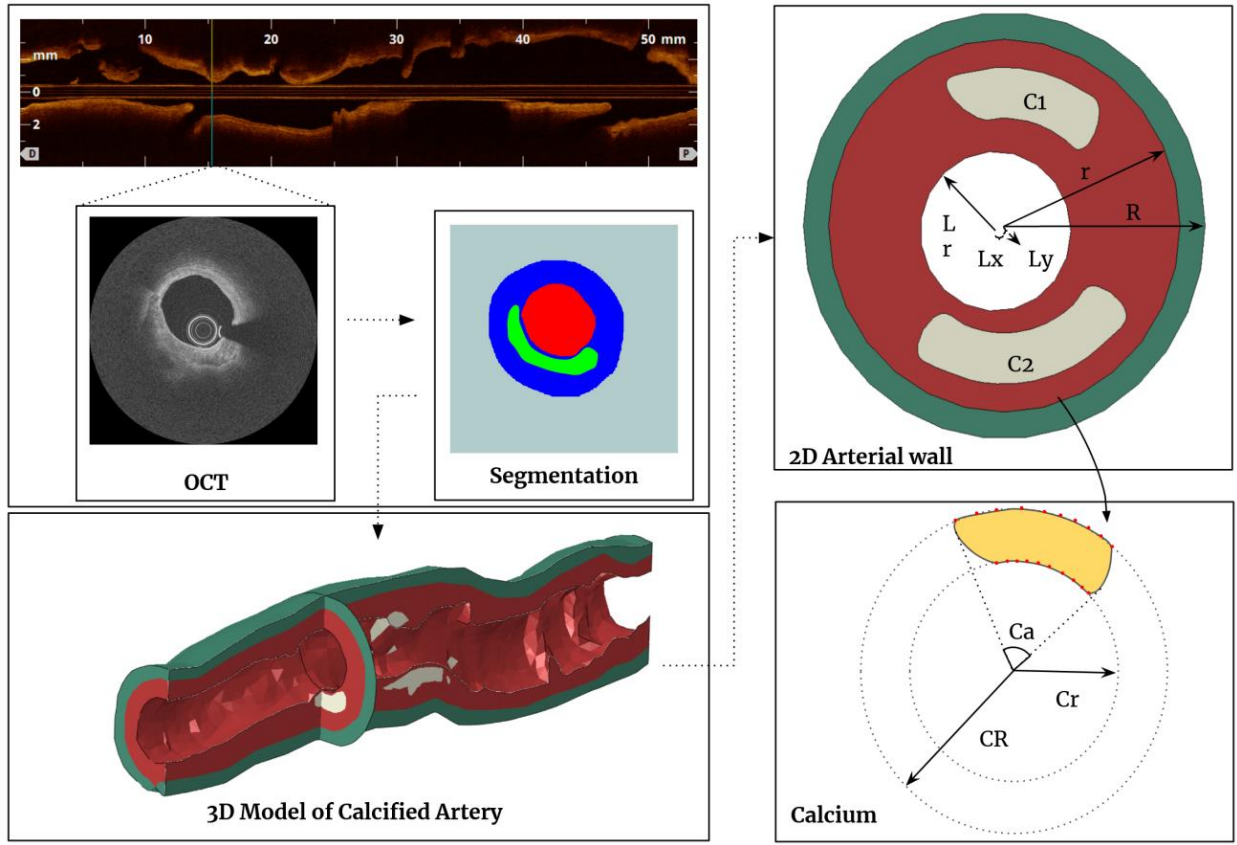

**Fig. 1** Using segmentation data, a 3D model of a calcified artery was constructed based on OCT data. To quickly generate various artery geometries, a 2D parametric model was developed using Python scripting in Abaqus/Explicit software. A simplified assumption of cross-sections consisting of circles and splines was made. Circle specifications, such as lumen, fibrous, artery, and calcium deposits, were randomly varied. Calcification was created by selecting two random circles as inner and outer boundaries, with a random angle between 0 and 180 degrees. Ten points were generated on inner and outer boundaries, and spline tools in Python were used to form a closed surface for calcification.

**Table 1** Features for random generation of idealized 2D artery geometry.

| Features | Symbol | Range |
| --- | --- | --- |
| Artery Outer Radius | R | 2 mm |
| Artery Inner Radius | r | 1.75 mm |
| Lumen Radius | Lr | 0.75 mm |

| | | | |
|---|---|---|---|
| Lumen Dislocation X | Lx | | -0.25-0.25 mm |
| Lumen Dislocation Y | Ly | | -0.25-0.25 mm |
| Number of Calcification | Cn | | 1-2 |
| Calcification Inner radius | Cr | | 1-1.25 mm |
| Calcification Outer radius | CR | | 1.3-1.5 mm |
| Calcification Angle | Ca | | 0-180° |

## 2.2. Finite Element Method Simulation Database

This study considers a two-dimensional plane strain cross-section of an arterial wall, depicted in Fig. 2, where the material is assumed to be hyperelastic (Table 2). The FE mesh used in the simulation is shown in Fig. 2, and a four-node constant strain element is discretized. To simulate the effect of blood pressure, a static pressure load of 140 mmHg (18.7 kPa)(Cilla et al., 2012) is applied to the boundary elements in the lumen, while nodes on the outside boundaries are restrained from displacement. The problem is solved using Abaqus software (Abaqus, 2011), and a typical converged stress and strain distribution is shown in Fig. 2.

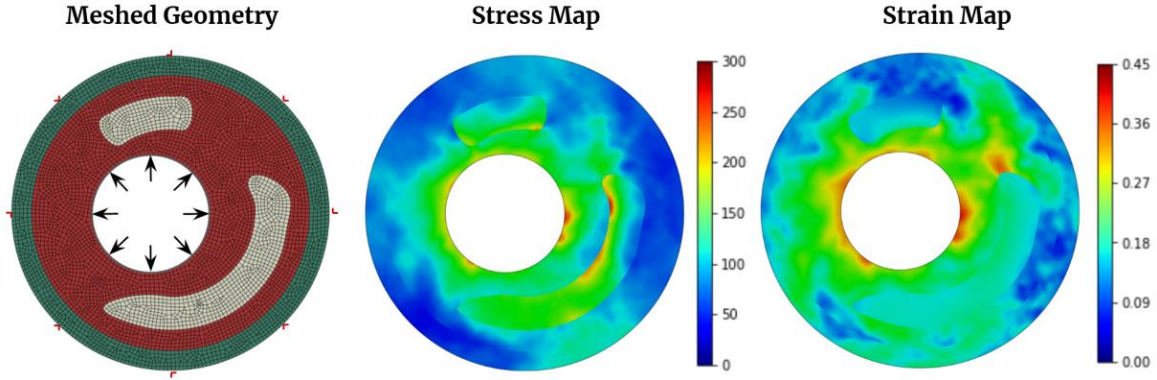

**Fig. 2** 2D arterial wall cross-section geometry meshed with four-node constant strain element and corresponding von Mises stress and strain distribution. The maximum stress and strain are considered 300 kPa and 0.45, respectively.

**Table 2** Material properties of arterial wall features (Dong et al., 2021).

| | $C_{10}$(MPa) | $C_{01}$(MPa) | $C_{11}$(MPa) | $C_{20}$(MPa) | $C_{02}$(MPa) | $C_{30}$(MPa) | $C_{03}$(MPa) |
|---|---|---|---|---|---|---|---|
| Artery | 0.108 | -0.101 | -0.179 | 0.088 | 0.062 | | |
| Fibrous | 0.040 | | | | 0.003 | | 0.0297 |
| Calcium | -0.495 | 0.506 | 1.193 | 3.637 | | 4.737 | |

## 2.3. Models and Methods
### 2.3.1. U-Net Architecture

By executing the FEM model on a range of arterial wall geometries, one can acquire a collection of input and output image data. These data can then be utilized to train a deep learning model in a supervised learning manner. An encoder-decoder network maps the input image to the output stress and strain map (Nie et al., 2020), depicted in Fig. 3(a). The binary maps provided as input depict the positions of calcium, lumen, and fibrous tissue. The encoder-decoder network converts these input images into a latent space with lower dimensions, which is subsequently mapped back to the stress and strain field. This approach assumes that the input and target spaces share a common latent space. The U-Net architecture (Ronneberger et al., 2015), based on the

FEM mapping data of images to stress and strain field, has been utilized for this purpose. The weights of architecture are trained through learning. Initially introduced in 2015 for medical image segmentation, the U-Net architecture is named after its U-shape, consisting of an encoder and a decoder. It has proven effective in capturing latent representations for various types of images. The encoder extracts feature from the input image using convolutional and max pooling layers.

In contrast, the decoder generates the output segmentation map using up-convolutional and concatenation layers. One of the significant advantages of the U-Net architecture is its ability to produce high-quality segmentation results even when trained on a limited amount of data. U-Net has been widely used in various medical image segmentation tasks, such as segmenting brain tumors, retinal blood vessels, and cell membranes. The architecture has also been extended and modified for other applications, such as semantic segmentation, instance segmentation, and object detection. The standard architecture in this study incorporates contracting and expanding layers along with skip connections to propagate context information and improve output resolution. The U-Net architecture utilized here is a slightly modified version of the original design. The encoder section consists of six repeating blocks, each containing a 2×2 max pooling operation, two consecutive 3×3 2D convolutions (except for the first and last blocks), batch normalization, and ReLU activation. The first decoder block employs a transpose convolution layer, while the last block consists of two consecutive convolution-batch norm-ReLU layers without a transpose convolution. The encoder and decoder blocks are followed by a final 1×1 convolutional layer, which reduces the 64-channel decoder output to a single channel. During weight training, the loss function is the weighted mean squared error (MSE) between the predicted and true von Mises stress and strain maps derived from the training data. The U-Net architecture was implemented using the TensorFlow and Keras libraries (Abadi et al., 2016). Fig. 3(b) illustrates a diagram of the modified U-Net architecture.

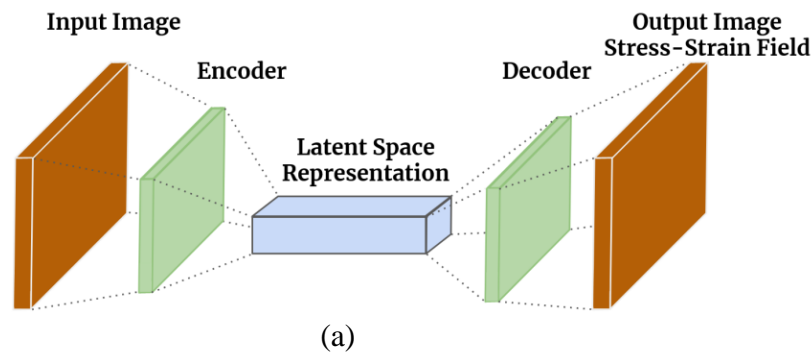

(a)

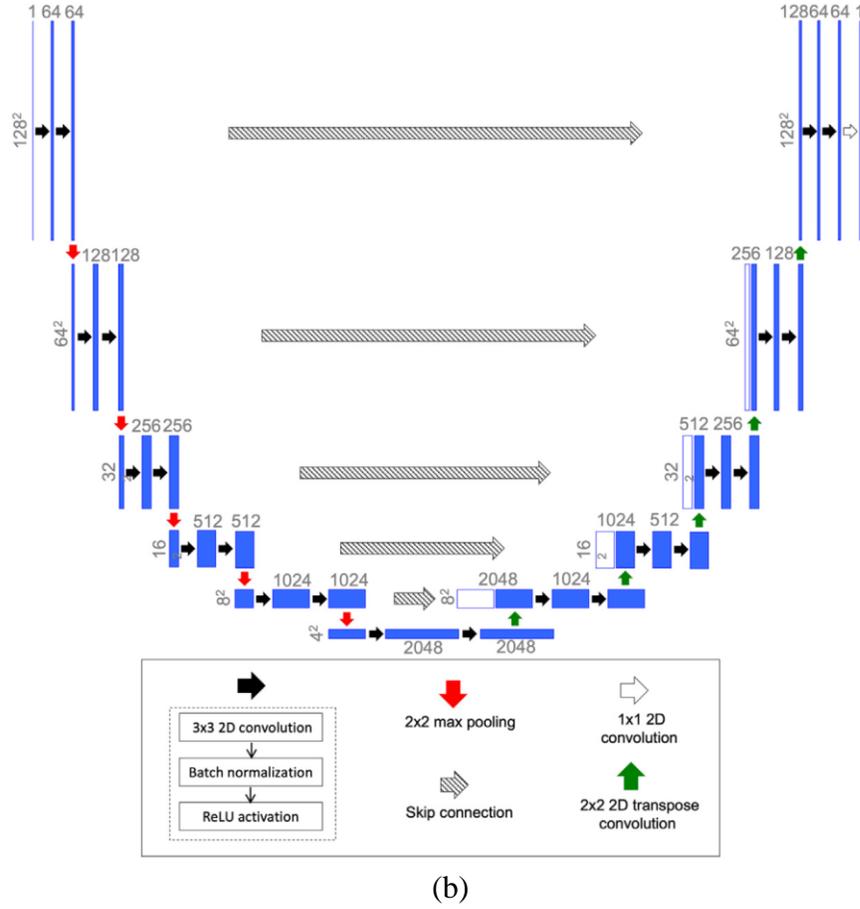

(b)

**Fig. 3 (a)** An encoder–decoder based network that can serve as an efficient surrogate for the FEM mapping and **(b)** the U-Net architecture(Ronneberger et al., 2015).

The training loss function is defined using the MSE, which is denoted as:

$$MSE = L[Y, f(X)] = \frac{1}{n} \sum_{i=1}^{n}[Y - f(X)]^2 \qquad (1)$$

Where X denotes the input image describing the 2D arterial wall images, n denotes the batch size, $f(X)$ indicates the prediction of the U-Net model, and Y denotes the output images. Performing verification to prevent overfitting during every training epoch also had been done in our network to select the best model without overfitting.

### 2.3.2. cGAN Architecture

The Generative Adversarial Network (GAN) is a deep neural network type that generates new data based on the data statistics of the training set (Creswell et al., 2018). GANs have two essential components, the generator, and the discriminator, which are trained against each other using game theory. The generator produces candidates that the discriminator evaluates. Though GANs were originally developed for unsupervised learning, incorporating labels as constraints can result in a subcategory known as conditional GANs (cGANs), as depicted in Fig. 4. Our research focuses on developing a deep learning model that utilizes a conditional generative adversarial network (cGAN) and paired image data (Isola et al., 2017). The cGAN model comprises two crucial components: the generator, known as U-Net, and the discriminator, referred to as

PatchGAN (Isola et al., 2017). The role of the generator is to take geometric images (labels or constraints) as input and produce field images of interest by incorporating random noise. Subsequently, the discriminator compares these generated field images to authentic images obtained from finite element modeling (FEM). The generator's objective is to increase the error rate of the discriminator.

In contrast, the discriminator aims to optimize its ability to distinguish between fake images generated by the generator and real ones. To accomplish this, we utilize TensorFlow (Abadi et al., 2016), a versatile machine learning framework, to perform calculations and implement the generative adversarial network (GAN) architecture for translating arterial wall geometries into strain fields (Isola et al., 2017). Our specific model employs U-Net as the generator and PatchGAN as the discriminator. U-Net generates synthetic strain field images based on arterial wall geometries, and PatchGAN evaluates the authenticity of these generated field images by comparing them to real strain field images. Our generator U-Net shares similarities with the CNN models discussed earlier, and the generator loss function is defined as:

$$Gen_{loss} = \lambda * gan_{loss} + L_1\_loss \qquad (2)$$

The variable gan_loss represents the sigmoid cross-entropy loss between the generated images and an array of ones. At the same time, L₁_loss calculates the mean absolute error between the generated and target images. It should be noted that (2) is equivalent to (1) when $\lambda$ is set to 0 and $L_1\_loss$ is replaced by $L_2\_loss$, which calculates the mean squared error. We tried several hyperparameters, such as $\lambda$ with different values ranging from 0.005 to 0.05. We saved the best models' weights to be imported to the new training model. We experimented with different architectures and numbers of layers with different filter sizes for the discriminator architecture and presented our final architecture.

The discriminator PatchGAN comprises five layers, which have around 11,000,000 trainable weights. The model receives two inputs and concatenates them using the concatenate layer. Afterward, the concatenated input is processed through five Conv2D layers with different filter sizes (64, 128, 256, 512, and 1024), a stride value of 2, and the same padding scheme. An activation function called LeakyReLU (He et al., 2015) is applied to the output tensors with an alpha value of 0.25. Finally, the last convolutional layer's output is sent to another Conv2D layer with a filter size of 1, a stride value of 1, and the same padding scheme, which generates a single output value. The $l_1$_regularizer is utilized as a kernel regularizer for each Conv2D layer, with a value of 1e-4. PatchGAN examines the generated field images by categorizing individual patches in the image as either real or fake. The Discriminator loss function is defined as

$$Dis\_loss = real\_loss + generated\_loss \qquad (3)$$

The variable real_loss represents the loss calculated using the sigmoid cross-entropy function (Yang et al., 2021) between real images and an array of ones. On the other hand, generated_loss represents the loss calculated using the sigmoid cross-entropy function between the generated images and an array of zeros. Our model is trained for 500 epochs, using a batch size of 4, to achieve convergence despite the common training instability observed in GANs. However, conditional GANs (cGANs) are more stable when the input labels are properly constrained. We carefully selected the number of training epochs to ensure that the generator learns sufficient information without the discriminator consistently classifying the generated images as fake

(Goodfellow, 2016). To determine the appropriate number of training epochs, we assessed the predictions of our ML model on the test set.

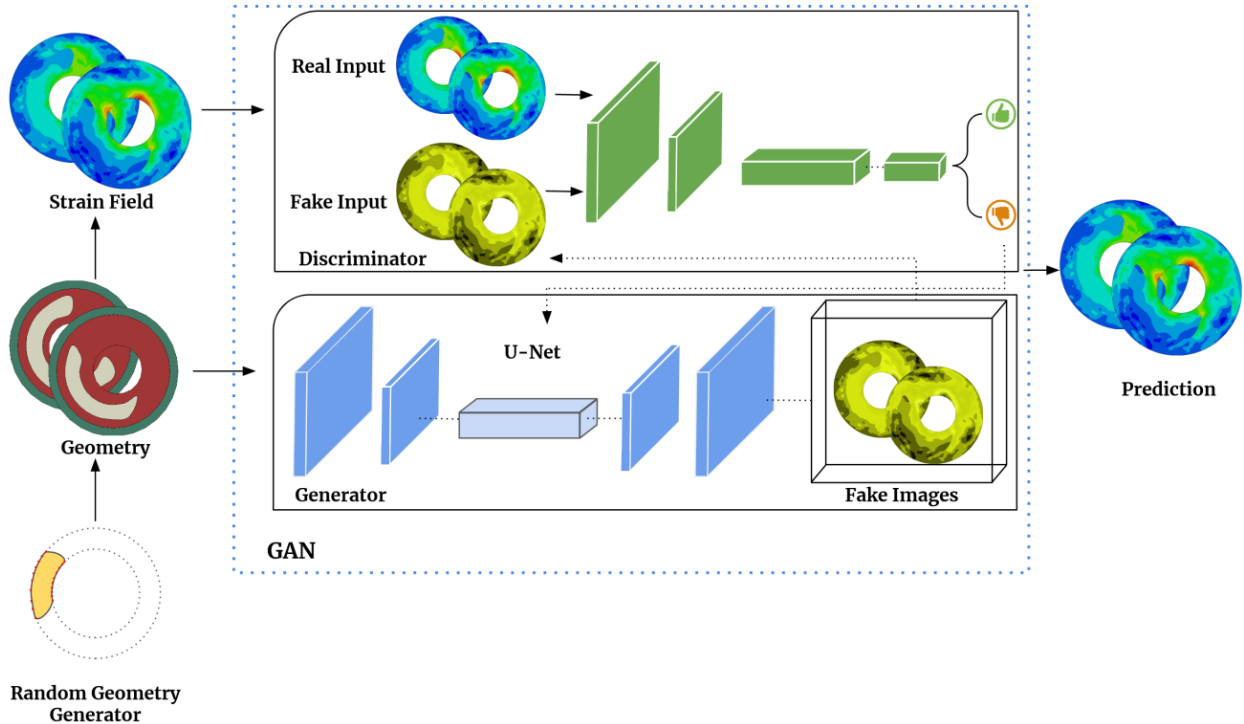

**Fig. 4** Workflow for the proposed method begins with a random generator using Python scripting to produce 2D arterial wall geometry images (256x256). Next, a FEM analysis is conducted to obtain arterial walls' true strain field information under blood pressure. ML model called Generative Adversarial Network (GAN), which includes a generator U-Net and a discriminator PatchGAN, is trained to predict strain fields from geometry images. The generator generates strain field maps using the geometry images as input. The discriminator then compares these generated images with real FEM-derived images. A well-trained model can accurately predict strain field maps, validated against high-fidelity FEM models and arterial walls with new geometries.

### 2.3.3. Ensemble learning

Combining multiple models in machine learning, known as ensemble learning (Dietterich, 2000), can significantly enhance predictive accuracy and robustness. This technique is especially effective when applied to U-Net and cGAN architectures. One method for implementing ensemble learning with U-Net/cGAN involves training various instances of the same architecture using different hyperparameters or random seeds. By averaging the predictions of these models, errors from any individual model can be minimized, leading to improved overall accuracy. Another approach explored in this paper involves combining different U-Net and cGAN generator architectures in an ensemble. An instance of an ensemble could comprise a conventional U-Net paired with a customized variation featuring supplementary skip connections. Alternatively, another ensemble may increase the number of filters in the convolutional layers of a standard U-Net and cGAN generator architectures by 32 channels. While this approach may improve feature extraction and result in more resilient predictions, it would also augment the model's parameter count and necessitate additional computational resources during training and inference. By capitalizing on the unique strengths of diverse U-Net and cGAN architectures through ensemble learning, it is frequently feasible to attain superior outcomes compared to solitary architecture.

This study employed two ensemble strategies for both U-Net and cGAN. The first strategy involved using the same U-Net architecture but with two different dropout rates in the layers - 0.2 and 0.35. After training the models with these rates, we averaged their results. This method was similarly applied to the cGAN generator architecture. The second strategy involved utilizing two different U-Net architectures, one with 10 layers and the other with 12 layers, as previously mentioned. We added a 32-channel layer at both the beginning and end of the network. Each model was trained with three dropout rates - 0.2, 0.3, and 0.35. In total, we trained 6 models using this approach. This same methodology was also applied to the cGAN architecture.

### 2.3.4. Deep transfer learning

Using transfer learning (Weiss et al., 2016), the previously learned weights of a deep learning model can be used as a starting point to retrain the same model on a new dataset, leading to faster convergence. To address the arterial walls problem, a combination of U-Net and cGAN architectures was utilized. The training process involved using 2D cross-section images along with their corresponding von Mises stress maps. This approach aimed to minimize the training requirements for a separate network that predicts strain maps. Because generating different output maps is costly, transfer learning was employed as a more efficient approach to predict strain maps. The von Mises stress trained model is utilized to predict the strain map.

### 2.4. Image quality metrics and statistical analysis

We used SSIM (Hore and Ziou, 2010) to measure the accuracy of our deep-learning models. In computer vision tasks, like image classification or object detection, the model's accuracy is often evaluated using metrics like precision, recall, and F1 score. However, these metrics do not provide information about the visual quality of the output images or the degree to which the model's predictions match the ground truth. SSIM can help fill this gap by measuring the structural similarity between the output and ground truth images. The output images from CNN can be compared to the ground truth images using SSIM, and the resulting value can be used to evaluate the model's accuracy. The range of SSIM is between 0 and 1, where a value of 1 indicates perfect similarity between two images, while a value of 0 indicates no similarity. The measure between two images x and y of common size N × N is:

$$SSIM(x,y) = \frac{(2\mu_x\mu_y + c_1)(2\sigma_{xy} + c_2)}{(\mu_x^2 + \mu_y^2 + c_1)(\sigma_x^2 + \sigma_y^2 + c_2)} \quad (4)$$

Where x and y are the two images being compared, $\mu_x$ and $\mu_y$ are the mean pixel intensities of images x and y, $\sigma_x^2$ and $\sigma_y^2$ are the variances of pixel intensities in images x and y, $\sigma_{xy}$ is the covariance of pixel intensities between x and y, and $c_1$ and $c_2$ are constants used to stabilize the division, typically $c_1 = (k_1 L)^2$ and $c_1 = (k_2 L)^2$, where $L$ is the dynamic range of the pixel values (for example, 255 for 8-bit grayscale images) and $k_1$ and $k_2$ are constants that control the relative contribution of the luminance and contrast terms to the final SSIM score. The SSIM formula considers the compared images' luminance, contrast, and structure and produces a value between 0 and 1.

## 3. Results
### 3.1. Stress map prediction accuracy

This section assesses the precision of matching 2D cross-section images of arterial walls with their corresponding von Mises stress map. To enhance the dataset, we leverage the physics of the problem by applying horizontal and/or vertical flipping to both the input images and output maps. This four-fold data augmentation is accomplished through three flipping operations: horizontal flip, vertical flip, and a combination of horizontal and vertical flip, as demonstrated in Fig. 5.

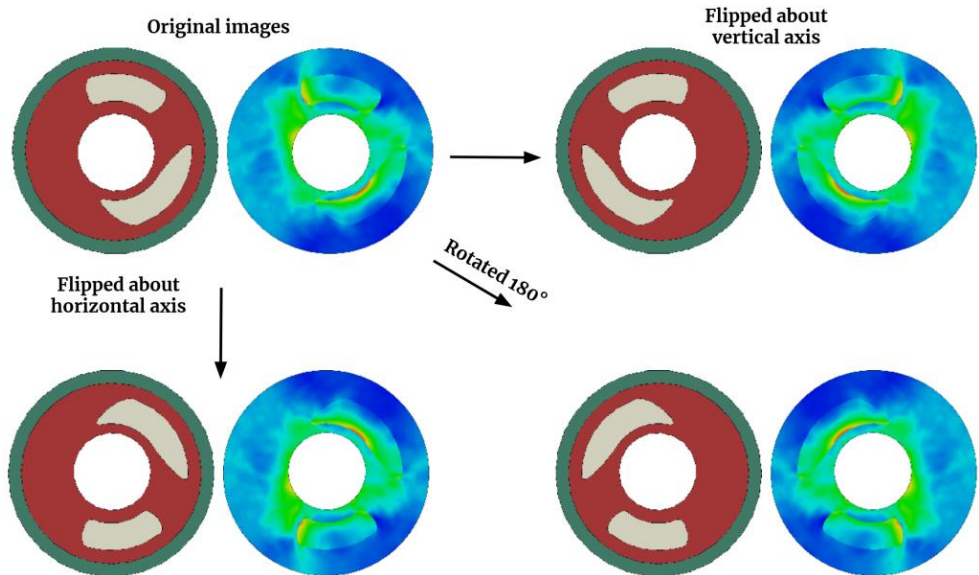

**Fig. 5** 4-fold data augmentation by image flipping.

After data augmentation, the training data consists of 4000 images of various spatial arrangements of calcium in the arterial wall and their corresponding von Mises stress maps for each model case (train: test 85%-15%). Fig. 6(a) and 6(b) displayed the actual (simulated via FEM) and predicted von Mises stress maps using U-Net and cGAN. In Fig. 6(a), a test data image is presented, followed by a comparison between the real stress map, calculated using FEM, and those generated by U-Net and cGAN. CGAN provides a superior prediction of von Mises stress compared to U-Net, achieving an SSIM of 0.837 and a mean error percentage of 6.80, whereas U-Net attains an SSIM of 0.831 and a mean error percentage of 8.3. Fig. 6(b) compares the stress map predictions made by the basic U-Net model and cGAN and the best U-Net and cGAN models obtained through ensemble techniques. The results for all models, including those using ensemble methods, are presented in Table 3. According to the table, the best ensemble model for U-Net was achieved by combining different U-Net architectures, as previously described. In this study, we utilized one U-Net model with 8 layers and another with 10 layers (by adding a layer with 32 channels at the beginning and end of the networks). The ensemble of these networks yielded the best results for U-Net, with a mean SSIM of 0.854. For cGAN, we applied the same technique to the generator layers and then combined it with a basic cGAN. As shown, the ensemble of two architectures also resulted in the best performance for cGAN (mean SSIM of 0.890). We then selected the best models and compared their predictions to those of the basic U-Net model and cGAN in Fig. 6(b). The ensemble technique for cGAN yielded superior predictions compared to the other models in this specific case, with an SSIM of 0.888.

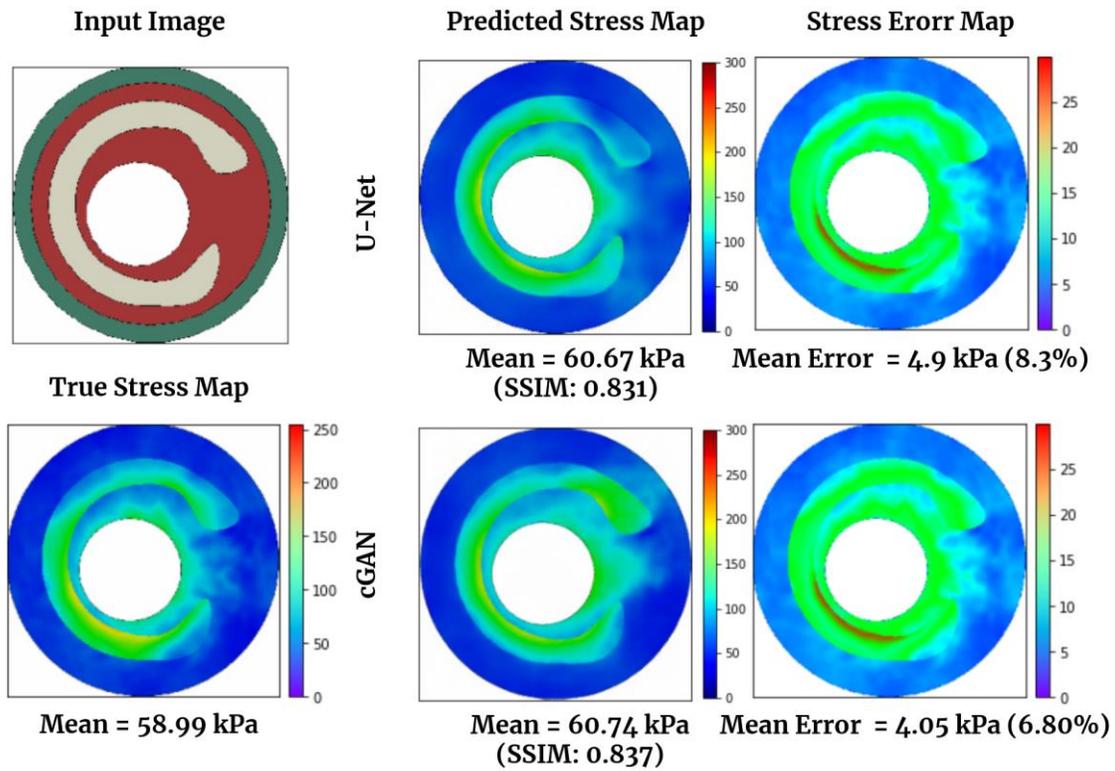

(a)

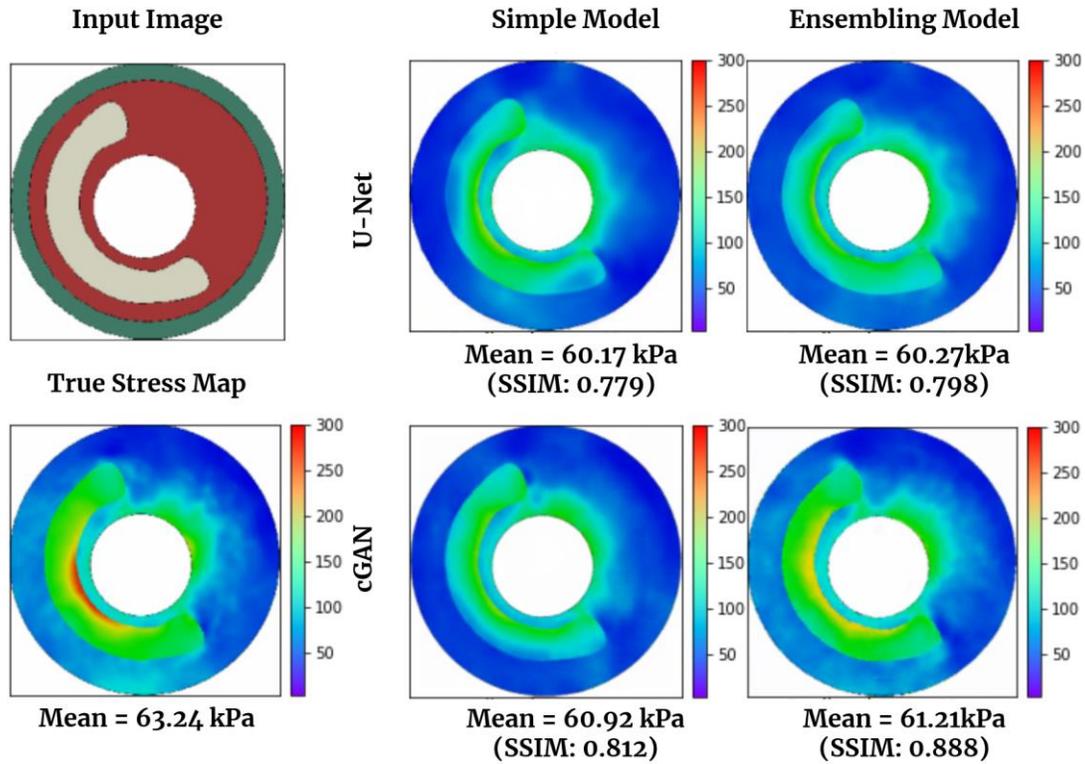

(b)

**Fig. 6** Von Mises stress map predicted from U-Net and cGAN architecture is based on 1000 FEM analyses of 2D arterial wall images, augmented to 4000 training images, **(a)** Test data image with true stress map (FEM) compared to U-Net and cGAN-generated maps. cGAN outperforms U-Net with SSIM 0.837 and 6.80% mean error. In comparison, U-Net has SSIM 0.831 and 8.3% mean error, **(b)** Comparison of stress map predictions between the best ensemble models and the basic U-Net and cGAN models, highlighting the superior performance of the ensemble cGAN (SSIM = 0.888).

**Table 3** Different approaches for predicting von Mises stress maps on the testing data.

|  | #of Models | Metrics | Min | Max | Mean |
|---|---|---|---|---|---|
| U-Net | 1 | MSE | 0.008 | 0.039 | 0.019 |
|  |  | SSIM | 0.773 | 0.902 | 0.842 |
| cGAN | 1 | MSE | 0.008 | 0.036 | 0.017 |
|  |  | SSIM | 0.787 | 0.895 | 0.850 |
| Ensembling (U-Net) | 2 | MSE | 0.008 | 0.038 | 0.018 |
|  |  | SSIM | 0.771 | 0.903 | 0.845 |
|  | 2×3 | MSE | 0.007 | 0.038 | 0.017 |
|  |  | SSIM | 0.778 | 0.906 | 0.854 |
| Ensembling (cGAN) | 2 | MSE | 0.005 | 0.023 | 0.010 |
|  |  | SSIM | 0.821 | 0.909 | 0.884 |
|  | 2×3 | MSE | 0.003 | 0.024 | **0.008** |
|  |  | SSIM | 0.823 | 0.924 | **0.890** |

Notably, the deep learning approach significantly reduces computational effort compared to FE results. While Abaqus requires approximately 2 minutes for a single analysis, our trained U-Net model predicts stress on a laptop CPU in less than a second. Our deep learning model has been trained on a GPU, specifically the NVIDIA GEFORCE RTX3090, within a Jupyter notebook environment. The utilization of this GPU significantly enhances computational speed compared to relying solely on a CPU. When running on GPU, each analysis takes approximately 0.06 seconds to complete.

### 3.2. Strain map prediction accuracy

This section evaluates the accuracy of mapping 2D cross-sectional images of arterial walls to their corresponding strain maps. Fig. 7(a) and 7(b) showcased the actual (simulated via FEM) and predicted strain maps using U-Net and cGAN. Fig. 7(a) displays a test data image, then compares the actual strain map (calculated using FEM) and the ones generated by U-Net and cGAN. U-Net outperforms cGAN in strain prediction, achieving an SSIM of 0.822 and a mean error percentage of 6.35, while cGAN achieves an SSIM of 0.803 and a mean error percentage of 5.34. Fig. 7(b) compares strain map predictions made by the basic U-Net model and cGAN and the best U-Net and cGAN models obtained through ensemble techniques. The results for all models, including those utilizing ensemble methods, are presented in Table 4. The best ensemble model for U-Net is achieved by combining different U-Net architectures, as previously described. We employed one U-Net model with 8 layers and another with 10 layers (adding a layer with 32 channels at the beginning and end of the networks). This ensemble yields the best results for U-Net, with a mean SSIM of 0.830. For cGAN, we applied the same technique to the generator layers and then combined it with a basic cGAN. This ensemble of two architectures also results in the best cGAN performance (mean SSIM of 0.803). We then chose the best models and compared their predictions to those of the basic U-Net model and cGAN in Fig. 7(b). The ensemble technique for U-Net produces superior predictions in this case, with an SSIM of 0.833. Based on the numerical results showing here and more we observed, though cGAN and its ensemble version did not deliver the best numbers in terms of SSIM, they do have the best performance visually.

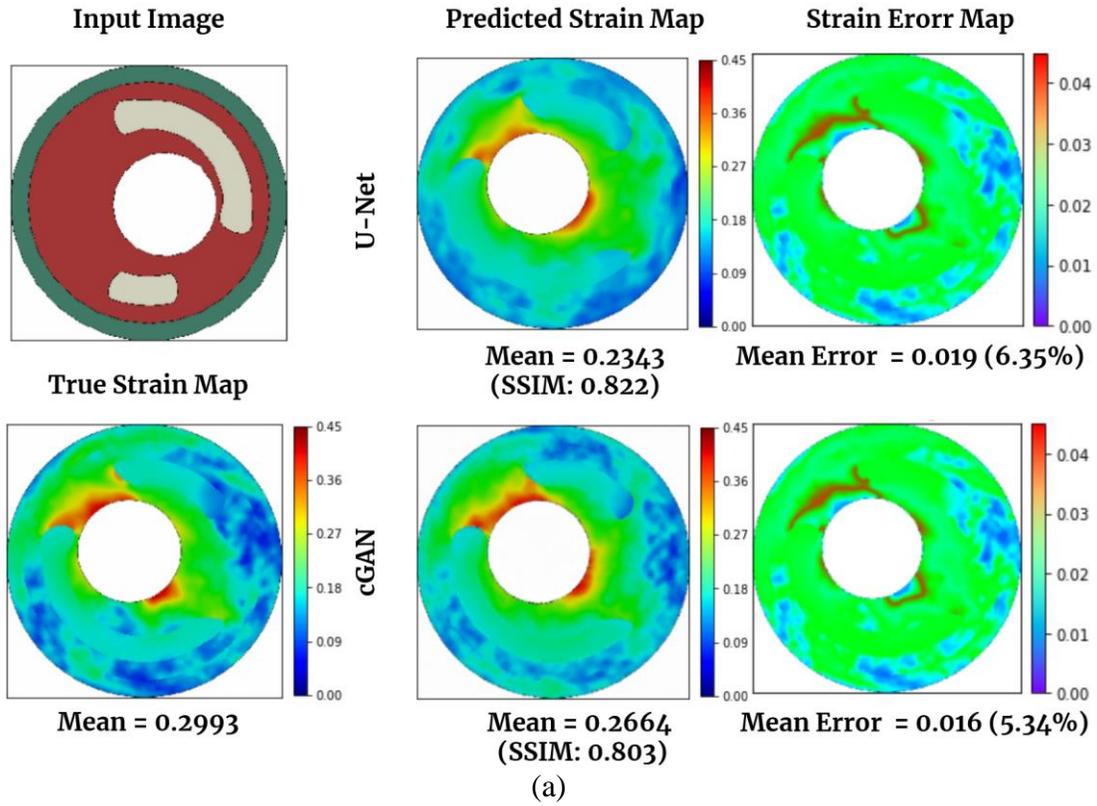

(a)

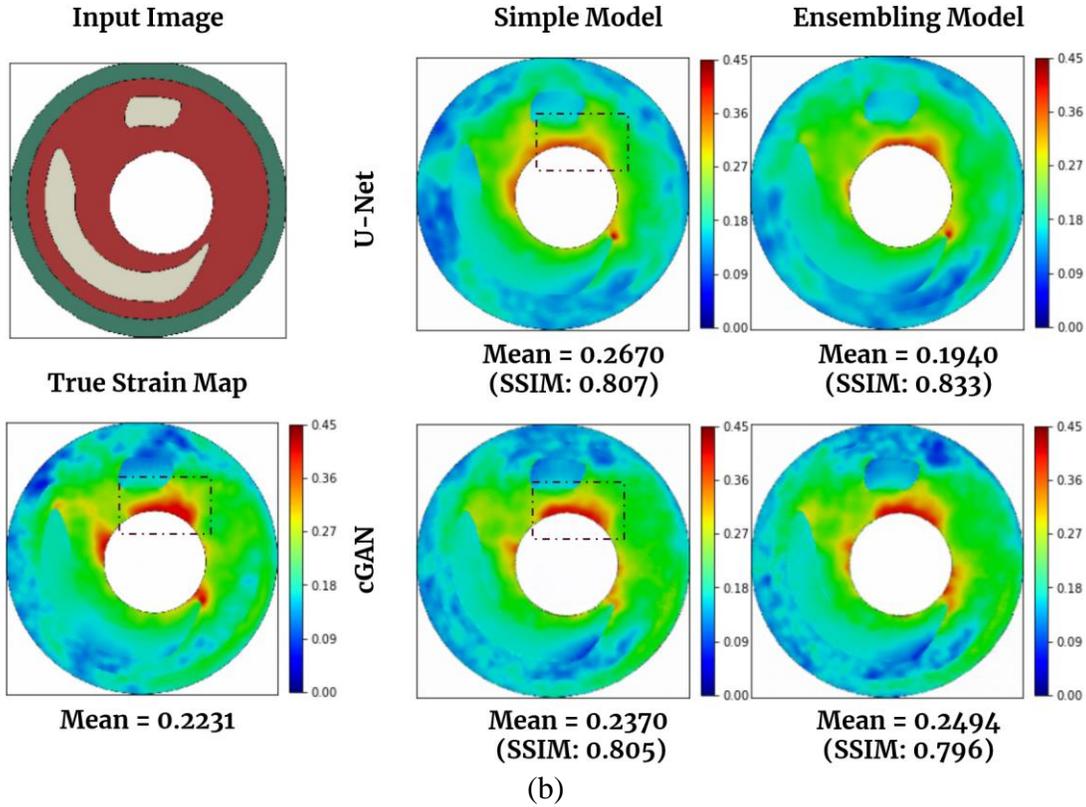

(b)

**Fig. 7** Strain map predicted from U-Net and cGAN architecture is based on 1000 FEM analyses of 2D arterial wall images, augmented to 4000 training images, **(a)** Test data image with true strain map (FEM) compared to U-Net and cGAN-generated maps. cGAN outperforms U-Net with SSIM 0.803 and 5.34% mean error. In comparison, U-Net has SSIM 0.822 and 6.35% mean error, **(b)** Comparison of strain map predictions between the best ensemble models and the basic U-Net and cGAN models, highlighting the superior performance of the ensemble U-Net (SSIM = 0.833), also, cGAN demonstrated superior performance in detecting higher strain values compared to U-Net, as highlighted by the black dashed box.

**Table 4** Different approaches for predicting strain maps on the testing data.

|  | #of Models | Metrics | Min | Max | Mean |
|---|---|---|---|---|---|
| U-Net | 1 | MSE | 0.0120 | 0.037 | 0.022 |
|  |  | SSIM | 0.750 | 0.866 | 0.814 |
| cGAN | 1 | MSE | 0.010 | 0.029 | 0.019 |
|  |  | SSIM | 0.751 | 0.842 | 0.793 |
| Ensembling (U-Net) | 2 | MSE | 0.009 | 0.034 | 0.019 |
|  |  | SSIM | 0.800 | 0.877 | 0.825 |
|  | 2×3 | MSE | 0.009 | 0.025 | 0.018 |
|  |  | SSIM | 0.8153 | 0.877 | **0.830** |
| Ensembling (cGAN) | 2 | MSE | 0.011 | 0.026 | 0.017 |
|  |  | SSIM | 0.785 | 0.839 | 0.800 |
|  | 2×3 | MSE | 0.011 | 0.021 | **0.017** |
|  |  | SSIM | 0.790 | 0.839 | 0.803 |
| Transfer Learning |  | MSE | 0.012 | 0.029 | 0.020 |

| (U-Net) | SSIM | 0.790 | 0.868 | 0.820 |
| Transfer Learning | MSE | 0.010 | 0.029 | 0.019 |
| (cGAN) | SSIM | 0.760 | 0.842 | 0.794 |

In Table 4, we display the SSIM and MSE values for the U-Net and cGAN models, which were trained using pre-trained weights from U-Net and cGAN models based on von Mises stress data. It was observed that both models led to an increase in SSIM and a decrease in MSE. Additionally, Fig. 8 compares the average SSIM and MSE values for stress and strain between the actual and predicted stress-strain maps across all 600 validation images, substantiating the high precision of our model's predictive capabilities. In Fig. 8(a), we compared the MSE of predicted von Mises stress maps among six networks: a simple U-Net, an ensemble method with various hyperparameters, an ensemble using two distinct U-Net architectures, a simple cGAN, an ensemble method with various hyperparameters in the cGAN generator model, an ensemble using two distinct cGAN generator architectures. The box plots reveal that employing an ensemble with two cGAN architectures results in a lower MSE than other models. Moreover, Fig. 8(b) indicates that the SSIM of an ensemble cGAN with two generator architectures is superior. Fig. 8(c) examines the MSE of eight strategies for predicting strain maps. A comparison of the U-Net and cGAN with a basic network, ensembles, and transfer learning reveals that an ensemble method with various hyperparameters in the cGAN generator model exhibits a narrower MSE range than other approaches. Concerning the SSIM of predicted strain maps, Fig. 8(d) illustrates that an ensemble using two distinct U-Net architectures outperforms other methods.

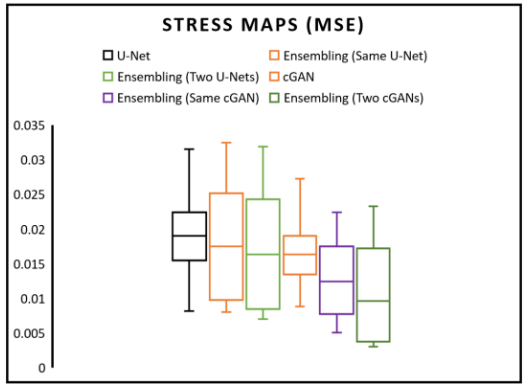
(a)

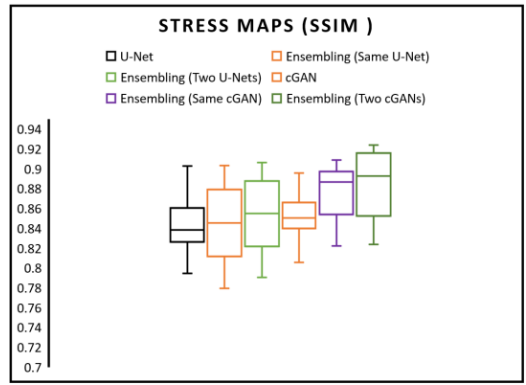
(b)

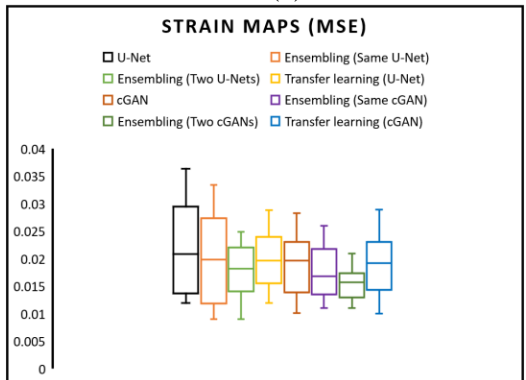
(c)

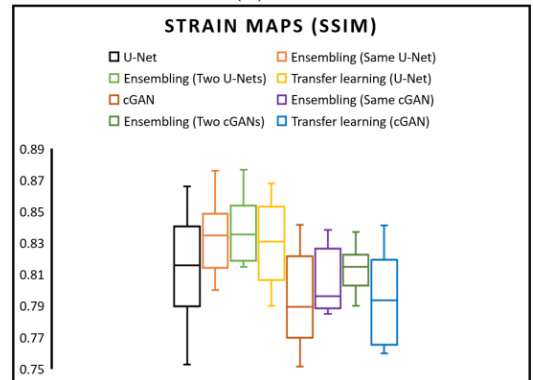
(d)

**Fig. 8** Comparison of MSE and SSIM for all 600 validation images, **(a)** MSE of all test data for prediction Stress maps using U-Net, Ensembling with same U-Net and different hyper parameters, and Ensembling with two U-Net architectures, cGAN, Ensembling with same cGAN generator model and different hyper parameters, and Ensembling with two cGAN generator architectures, **(b)** SSIM of all test data for prediction Stress maps using U-Net, Ensembling with same U-Net and different hyper parameters, and Ensembling with two U-Net architectures, cGAN, Ensembling with same cGAN generator model and different hyper parameters, and Ensembling with two cGAN generator architectures, **(c)** MSE of all test data for prediction Strain maps using U-Net, Ensembling with same U-Net and different hyper parameters, Ensembling with two U-Net architectures, U-Net using transfer learning, cGAN, Ensembling with same cGAN generator and different hyper parameters, Ensembling with two cGAN generator architectures, and cGAN using transfer learning **(d)** SSIM of all test data for prediction Strain maps using U-Net, Ensembling with same U-Net and different hyper parameters, Ensembling with two U-Net architectures, U-Net using transfer learning, cGAN, Ensembling with same cGAN generator and different hyper parameters, Ensembling with two cGAN generator architectures, and cGAN using transfer learning.

## 4. Discussion and Conclusion

In summary, this research aims to explore the potential of deep learning tools in predicting stress-strain fields within 2D cross sections of arterial wall geometries, which can replace the traditional FEM method. To establish a relationship between the spatial distribution of calcification in arterial wall cross-sections and the von Mises stress and strain fields, two ML networks were investigated. The first network utilized a U-Net architecture within a CNN model, while the second employed a cGAN. The trained models can accurately predict stress-strain field maps for arterial walls with varying quantities and spatial configurations of calcification, which is a significant achievement. To generate the training dataset, a Python code was utilized to create various shapes of arterial walls with randomized features such as location, angle, thickness, number of calcifications, lumen area location, and diameters, drawing from the simplified clinical data and our observations. The dataset consists of input and output images obtained by applying boundary conditions and extracting stress-strain field maps. The trained U-Net models produce precise predictions of the von Mises stress and strain fields, with SSIM of 0.854 and 0.830 and MSE of 0.017 and 0.018 for stress and strain, respectively, on a held-out test set. Additionally, the stress and strain data-trained cGAN model outperforms the CNN model in prediction accuracy, demonstrating SSIM of 0.890 and 0.803 and MSE of 0.008 and 0.017 for stress and strain, respectively. However, the utilized image quality metrics have limitations. They cannot encompass all the image perception details significant for human visual systems. We expect the performance of cGAN can be evaluated more accurately and fairly if a comprehensive metric could be designed in future.

This research demonstrates the ability of a surrogate model for finite element analysis to accurately predict stress-strain fields of arterial walls, utilizing 2D cross-sectional images. Furthermore, we showcased the numerous benefits of utilizing ensemble and transfer learning methodologies to achieve superior levels of precision. By combining multiple models, an ensemble approach can significantly enhance predictive performance, as the diverse models can compensate for each other's weaknesses and amplify their strengths. On the other hand, transfer learning can leverage the knowledge gained from a pre-trained model and apply it to a new task with a limited amount of data. This approach can drastically reduce the required training and improve the model's performance on the new task. Our findings indicate that by ensembling the same U-Net with various hyperparameters, the mean SSIM across all testing data can be boosted from 0.814 to 0.825. Moreover, by incorporating an additional layer into the standard U-Net and training two models thrice, we achieved the highest SSIM of 0.830 for predicting the strain map. Transfer learning was utilized in addition to the von Mises stress trained model to forecast strain maps. By comparing the loss of a U-Net model trained from scratch to that of a pretrained U-Net model trained with von Mises stress data over 100 epochs, it was noted that both models displayed convergence and exhibited similar behavior during training. Despite this, the transfer learning model demonstrated quicker convergence and produced an average SSIM of 0.820 for all test data, outperforming a U-Net trained from scratch. We employed a cGAN for stress and strain map predictions and compared the outcomes with those derived from a U-Net. The optimal stress map predictions were attained by assembling two generator architectures in the cGAN, achieving an SSIM of 0.890 (Table 3). Despite this, U-Net demonstrated superior performance in strain prediction, obtaining the highest SSIM of 0.830 when combined with two U-Net architectures. A comparison of the SSIM and MSE for cGANs trained with transfer learning and those trained from scratch revealed that transfer learning could marginally enhance SSIM from 0.793 to 0.794 while maintaining the same MSE. By incorporating ensemble and transfer learning techniques, our experiments demonstrated that we could achieve exceptional results even in scenarios with limited data and challenging problems.

In upcoming studies, transfer learning could predict additional properties, such as displacement or distribution maps of arterial walls since they share comparable distribution color patterns with stress and strain maps. Although this study focused on two-dimensional situations, the methodology presented can be extended to predict stress and strain distribution in arterial walls by utilizing a three-dimensional variant for CNN training and testing. Additionally, the proposed approach can estimate other arterial wall properties. It is worth noting that this work did not incorporate lipids or other tissues in the arterial wall geometry. However, these can be easily included using Python scripting in Abaqus. Expanding the model to include multiple tissues and exploring additional properties, such as arterial wall displacement and deformation, would be valuable for future work. Another potential area of investigation could be incorporating stenting in the simulation to predict arterial wall deformation.

**Data Availability Statement:**

The training data for the study is available upon request.

**Declaration of competing interest:**

The authors declare that they have no known competing financial interests or personal relationships that could have appeared to influence the work reported in this paper